%% file: fusion14_camera.tex
\documentclass[english,letter]{IEEEtran}
\usepackage[T1]{fontenc}
\usepackage[latin9]{inputenc}
\usepackage{float}
\usepackage{calc}
\usepackage{amsthm}
\usepackage{amsmath}
\usepackage{amssymb}
\usepackage{graphicx}
\usepackage{esint}
\usepackage{xargs}[2008/03/08]

\makeatletter

\floatstyle{ruled}
\newfloat{algorithm}{tbp}{loa}
\providecommand{\algorithmname}{Algorithm}
\floatname{algorithm}{\protect\algorithmname}

\theoremstyle{plain}
\newtheorem{thm}{\protect\theoremname}
\theoremstyle{plain}
\newtheorem{prop}[thm]{\protect\propositionname}

\usepackage{flushend}

\makeatother

\usepackage{babel}
\providecommand{\propositionname}{Proposition}
\providecommand{\theoremname}{Theorem}

\begin{document}

\title{A Random Finite Set Model for Data Clustering}

\author{Dinh Phung$^{\dagger}$ and Ba-Ngu Vo$^{\ddagger}$\thanks{Acknowledgement: The work of the 2nd author is supported by the Australian Research Council under the Future Fellowship FT0991854.}\\
$^{\dagger}$Center for Pattern Recognition and Data Analytics, Deakin
University, Australia\\
$^{\ddagger}$Department of Electrical and Computer Engineering, Curtin
University, Australia\\
Email\emph{: dinh.phung@deakin.edu.au, ba-ngu.vo@curtin.edu.au}}
\maketitle
\begin{abstract}
The goal of data clustering is to partition data points into groups
to optimize a given objective function. While most existing clustering
algorithms treat each data point as vector, in many applications each
datum is not a vector but a point pattern or a set of points. Moreover,
many existing clustering methods require the user to specify the number
of clusters, which is not available in advance. This paper proposes
a new class of models for data clustering that addresses set-valued
data as well as unknown number of clusters, using a Dirichlet Process
mixture of Poisson random finite sets. We also develop an efficient
Markov Chain Monte Carlo posterior inference technique that can learn
the number of clusters and mixture parameters automatically from the
data. Numerical studies are presented to demonstrate the salient features
of this new model, in particular its capacity to discover extremely
unbalanced clusters in data.

\input{macros.tex}

\end{abstract}

\section{Introduction}

Stochastic geometry is an established area of study with a long history
that dates back to the famous problem of Buffon's needle \cite{Stoyan95}.
Stochastic geometric models, including deformable templates and random
finite sets have long been used by statisticians to develop techniques
for object recognition in static images \cite{Baddeley_Lieshout_93}.
Random finite set (RFS) theory (or more generally point process theory)
is the study of random point patterns with applications spanning numerous
disciplines from agriculture/forestry and epidemiology/public health
\cite{Stoyan_Penttien_00}, \cite{Moller04} to communications \cite{baccelli_bk10},
target tracking \cite{Mahler_03,Vo_etal_05}, computer vision \cite{Baddeley_Lieshout_93},
and robotics \cite{MSLAM2}. The common theme in these applications
is the set-valued observation and/or set-valued parameters.

While RFS theory is suitable for inferencing problems involving unknown
and random number of parameters, its use has been largely overlooked
in \emph{the problem of learning from data}. One of the most popular
tasks in learning from data is data clustering where the goal is to
partition the data points into groups to optimize a given objective
function, such as the distance between data points within a group
as in the K-means algorithm. Many clustering methods require the number
of clusters to be known apriori, but this is not the case in practice.
Nearly all existing clustering algorithms treat each data point as
a vector. However, in many applications each data point is a set of
vectors (rather than a vector of fixed dimension). For example, in
image analysis, the information content of an image is summarized
and stored as a set of features. Another example is text modelling,
where the `bag-of-words' representation treats a document as a finite
set of words, since the order of appearance of the words is neglected.
Other examples include geo-spatial data, epidemiological data etc.
In general, a sparse data point in which the order of the non-zero
elements is not important can be represented as a set-valued data
point.

In this paper we propose a new class of model for data clustering
that addresses set-valued data as well as unknown number of clusters
based on Poisson RFS. The proposed model is a Dirichlet process mixture
of Poisson RFS and is termed the \emph{Dirichlet}\textbf{\emph{ }}\emph{Poisson
RFS Mixture Model} (DP-RFS). In particular, we derive a family of
conjugate priors for Poisson RFS likelihoods, and use this result
to develop an infinite mixture of Poisson RFS likelihoods with Dirichlet
process prior on the mixture weights. We then present an efficient
Markov Chain Monte Carlo method to perform posterior inference, from
which the number clusters and mixture parameters are automatically
learned from the data. More specifically, we exploit the conjugacy
of the prior on the parameters of the Poisson RFS likelihood to integrate
over these parameters and derive an efficient collapsed Gibbs sampler
that converges faster than a standard full Gibbs sampler. A numerical
study is presented to demonstrate the capability of the proposed DP-RFS
model to learn in scenarios with extremely unbalanced clusters where
existing methods typically fail.

\section{Background}

\subsection{Finite Bayesian mixture models\label{sub:Finite-Bayesian-mixture}}

The most common probabilistic approach to clustering is mixture modelling
where the clustering process is treated as a density estimation problem.
Mixture models assume in advance the existence of $K$ latent subpopulations
in the data and specifies a likelihood of observing each data point
$x$ as a mixture: 
\begin{align}
p\left(x\gv\pi_{1:K},\phi_{1:K}\right) & =\sum_{k=1}^{K}\pi_{k}f\left(x\gv\phi_{k}\right)\label{eq:mm_lik-1}
\end{align}
where $\pi_{k}$ is the probability that $x$ belongs to the $k$-th
sub-population and $\sum_{k=1}^{K}\pi_{k}=1$. This is the parametric
and frequentist approach to mixture modeling. The EM algorithm is
typically employed to estimate the parameters $\pi_{1:K}$ and $\phi_{1:K}$
from the data. Gaussian mixture models (GMM), for instance, is commonly
used in signal processing and target tracking. In this case, each
mixture-specific parameter $\phi_{k}$ consists of $\left(\bmu_{k},\Sigma_{k}\right)$
which specifies the mean and covariance matrix for each mixture.

Under a Bayesian setting \cite{Gelman_etal_bk03_Bayesian,Robert_bk01}
the parameters $\pi_{1:K}$ and $\phi_{1:K}$ are further endowed
with suitable prior distributions. Typically a symmetric Dirichlet
distribution $\dirpdf\left(\cdot\gv\eta\right)$ is used as the prior
of $\pi_{1:K}$, while the prior distribution for $\phi_{1:K}$ is
model-specific depending on the form of the likelihood function $f$
which admits a conjugate prior $h$.  A Bayesian mixture model specifies
the generative likelihood for $x$ as:
\begin{align*}
p\left(x\gv\eta,h\right) & =\int\int\sum_{k=1}^{K}\pi_{k}f\left(x\gv\phi_{k}\right)P(d\pi_{1:K})P\left(d\phi_{1:K}\right)
\end{align*}
Under this formalism, inference amounts to deriving the joint posterior
distribution for $\pi_{1:K}$ and $\phi_{1:K}$, which is often intractable.
Markov Chain Monte Carlo methods, such as Gibbs sampling, are common
approaches for the inference task \cite{Gelman_etal_bk03_Bayesian,bernardo2009bayesian}. 

Suppose there are data points $D=\{x_{1},...,x_{N}\}$. A latent indicator
variable $z_{i}$ is introduced for each data point $x_{i}$ to specify
its mixture component where $z_{i}\in\{1,...,K\}$ and $\Pr\left(z_{i}=k\right)=\pi_{k}$.
Conditioning on this latent variable, the distribution for $x_{i}$
simplifies to:
\begin{align}
p\left(x_{i}\gv z_{i}=k,\phi_{1:K},\pi_{1:K}\right) & =f\left(x_{i}\gv\phi_{k}\right)\label{eq:-1}
\end{align}
Full Gibbs sampling for posterior inference becomes straightforward
by iteratively sampling the conditional distributions among the latent
variables $\pi_{1:K}$, $z_{i}$ and $\phi_{k}$, i.e.,
\begin{align}
p\left(z_{i}\gv z_{-i},x_{1:n},\pi_{1:K},\phi_{1:K}\right) & \propto p\left(x_{i}\gv z_{i}\right)=f\left(x_{i}\gv\phi_{z_{i}}\right)\label{eq:full_Gibbs1}\\
p\left(\pi_{1:K}\gv z_{1:n},x_{1:n},\phi_{1:K}\right) & \propto p\left(z_{1:n}\gv\pi_{1:K}\right)p\left(\pi_{1:K}\right)\label{eq:full_Gibbs2}\\
p\left(\phi_{k}\gv z_{1:n},x_{1:n},\pi_{1:K},\phi_{-k}\right) & \propto\prod_{x\in\mathfrak{X}_{k}}p\left(x\gv\phi_{k}\right)p\left(\phi_{k}\right)\label{eq:full_Gibbs3}
\end{align}
where $\mathfrak{X}_{k}=\left\{ x_{i}:z_{i}=k,i=1,\ldots N\right\} $
is the set of all data points assigned to component $k$, and $z{}_{-i}$
denotes the set of all assignment indicators except $z_{i}$, i.e.,
$z_{-i}=(z,\ldots,z_{i-1},z_{i+1},...,z_{N})$. Due to the conjugacy
of Multinomial and Dirichlet distributions the posterior for $\pi_{1:K}$
is again a Dirichlet; and with a conjugate prior, the posterior for
$\phi_{1:K}$ will remain in the same form, hence they are straightforward
to sample. Collapsed Gibbs inference scheme can also be developed
to improve the variance of the estimators by integrating out $\pi_{1:K}$
and $\phi_{1:K}$ , leaving out the only following conditional to
sample from:

$p\left(z_{i}=k\gv z_{-i},x_{1:N}\right)$
\begin{equation}
\propto\frac{\int p\left(x_{i}\gv\phi_{k}\right)p\left(\phi_{k}\gv\left\{ x_{j}:j\neq i,z_{j}=k\right\} \right)d\phi_{k}}{\eta+n_{-i,k}}
\end{equation}
where $\eta$ is the hyperparameter for $\pi_{1:K}$, assumed to be
a symmetric Dirichlet distribution, and ${\textstyle n_{-i,k}=\sum_{j=1,j\neq i}^{N}}1_{z_{j}}(k)$
is the number of assignments to cluster $k$, excluding position $i$.
The second term involves an integration which can easily be recognized
as the predictive likelihood under the posterior distribution for
$\phi_{1:K}$. For conjugate prior, this expression can be analytically
evaluated. Several results can readily be found in many standard Bayesian
text book such as \cite{Gelman_etal_bk03_Bayesian}.

A key theoretical limitation in the parametric Bayesian mixture model
described so far is the assumption that the number of mixtures $K$
in the data is known and \emph{one has to specify it in advance} to
apply this model. Recent advances in Bayesian nonparametric modeling
(BNP) (e.g., see \cite{Ghosh_bk03_Bayesian,Hjort_etal_bk10_bayesian})
provides a principled alternative to overcome these problems by introducing
a nonparametric prior distribution on the parameters, which can be
derived from Poisson point process or RFS.

\subsection{Poisson RFS}

The Poisson RFS, which models ``no interaction\textquotedbl{} or
``complete spatial randomness\textquotedbl{} in spatial point patterns,
is arguably one of the best known and most tractable of point processes
\cite{Stoyan95,Daley88,vanLieshout00,Moller04,Kingman93}. The Poisson
RFS itself arises in forestry \cite{Stoyan_Penttien_00}, geology
\cite{Ogata_99}, biology \cite{marmarelis2005general}, particle
physics \cite{marmarelis2005general}, communication networks \cite{baccelli_bk10},
\cite{haenggi2005distances}, \cite{haenggi2009stochastic} and signal
processing \cite{Mahler_03}, \cite{SVBZ}, \cite{caron2011conditional}.
The role of the Poisson RFS in point process theory, in most respects,
is analogous to that of the normal distribution in random vectors
\cite{Cox_Isham_bk80}. 

We briefly summarize the concept of Poisson RFS since this is needed
to address the problem of unknown number of clusters and set-valued
data. An RFS $X$ on a state space $\mathcal{X}$ is random variable
taking values in $\mathcal{F}(\mathcal{X})$, the space of finite
subsets of $\mathcal{X}$. RFS theory is a special case of point process
theory--the study of random counting measures. An RFS can be regarded
as a simple-finite point process, but has a more intuitive geometric
interpretation. For detailed treatments, textbooks such as \cite{Stoyan95,Daley88,vanLieshout00,Moller04}.

Let $|X|$ denotes the number of elements in a set $X$ and $\left\langle f,g\right\rangle =\int f\left(x\right)g\left(x\right)dx$.
An RFS $X$ on $\mathcal{X}$ is said to be \emph{Poisson} with a
given \emph{intensity function} $v$ (defined on $\mathcal{X}$) if
\cite{Stoyan95,Daley88}:
\begin{enumerate}
\item for any $B\subseteq\mathcal{X}$ such that $\left\langle v,1_{B}\right\rangle <\infty$,
the random variable $|X\cap B|$ is Poisson distributed with mean
$\left\langle v,1_{B}\right\rangle $,
\item for any disjoint $B_{1},...,B_{i}\subseteq\mathcal{X}$, the random
variables $|X\cap B_{1}|,...,|X\cap B_{i}|$ are independent. 
\end{enumerate}
Since $\left\langle v,1_{B}\right\rangle $ is the expected number
of points of $X$ in the region $B$, the intensity value $v(x)$
can be interpreted as the instantaneous expected number of points
per unit hyper-volume at $x$. Consequently, $v(x)$ is not dimensionless
in general. If hyper-volume (on $\mathcal{X}$) is measured in units
of $\mathcal{K}$ (e.g. $m^{d}$, $cm^{d}$, in$^{d}$, etc.) then
the intensity function $v$ has unit $\mathcal{K}^{-1}$.

The number of points of a Poisson point process $X$ is Poisson distributed
with mean $\left\langle v,1\right\rangle $, and condition on the
number of points the elements $x$ of $X$ are independently and identically
distributed (i.i.d.) according to the probability density $v(\cdot)/\left\langle v,1\right\rangle $
\cite{Stoyan95,Daley88,vanLieshout00,Moller04}. It is implicit that
$\left\langle v,1\right\rangle $ is finite since we only consider
simple-finite point processes.

The probability distribution of a Poisson point process $X$ with
intensity function $v$ is given by (\cite{Moller04} pp. 15):
\begin{align}
 & \Pr(X\in\mathcal{T})\nonumber \\
 & \,\,\,=\sum_{i=0}^{\infty}\frac{e^{-\left\langle v,1\right\rangle }}{i!}\int_{\mathcal{X}^{i}}\!\!1_{\mathcal{T}}(\{x_{1},...,x_{i}\})v^{\{x_{1},...,x_{i}\}}d(x_{1},...,x_{i})\label{eq:Poissonmeas}
\end{align}
for any (measurable) subset $\mathcal{T}$ of $\mathcal{F}(\mathcal{X})$,
where $\mathcal{X}^{i}$ denotes an $i$-fold Cartesian product of
$\mathcal{X}$, with the convention $\mathcal{X}^{0}=\{\emptyset\}$,
the integral over $\mathcal{X}^{0}$ is $1_{\mathcal{T}}(\emptyset)$
and $v^{X}=\prod_{x\in X}v\left(x\right)$. A Poisson point process
is completely characterized by its intensity function (or more generally
the intensity measure).

Probability densities of random finite sets considered in this work
are defined with respect to the reference measure $\mu$ given by
\begin{equation}
\mu(\mathcal{T})=\sum_{i=0}^{\infty}\frac{1}{i!\mathcal{K}^{i}}\int_{\mathcal{X}^{i}}1_{\mathcal{T}}(\{x_{1},...,x_{i}\})d(x_{1},...,x_{i})\label{eq:referencemeasure}
\end{equation}
for any (measurable) subset $\mathcal{T}$ of $\mathcal{F}(\mathcal{X})$.
The measure $\mu$ is analogous to the Lebesque measure on $\mathcal{X}$
(indeed it is the unnormalized distribution of a Poisson point process
with unit intensity $v=1/\mathcal{K}$ when the state space $\mathcal{X}$
is bounded). Moreover, it was shown in \cite{Vo_etal_05} that for
this choice of reference measure, the integral of a function $f:\mathcal{F}(\mathcal{X})\rightarrow\mathbb{R}$,
given by
\begin{equation}
\int\!\! f(X)\mu(dX)=\sum_{i=0}^{\infty}\frac{1}{i!\mathcal{K}^{i}}\!\!\int_{\mathcal{X}^{i}}\! f(\{x_{1},...,x_{i}\})d(x_{1},...,x_{i}),\label{eq:integral}
\end{equation}
is equivalent to Mahler's set integral \cite{Mahler_03}. Note that
the reference measure $\mu$, and the integrand $f$ are all dimensionless.
Probability densities for Poisson RFS take the form:
\begin{equation}
f(X)=\mathcal{K}^{|X|}e^{-\left\langle u,1\right\rangle }u^{X}.\label{eq:Poissondensity}
\end{equation}
Note that for any (measurable) subset $\mathcal{T}$ of $\mathcal{F}(\mathcal{X})$
\begin{eqnarray*}
 &  & \!\!\!\!\!\!\!\!\!\!\!\!\!\!\!\!\!\!\!\int_{\mathcal{T}}f(X)\mu(dX)\\
 & = & \!\!\!\int1_{\mathcal{T}}(X)f(X)\mu(dX)\\
 & = & \!\!\!\sum_{i=0}^{\infty}\!\frac{e^{-\left\langle u,1\right\rangle }}{i!}\!\!\int_{\mathcal{X}^{i}}\!\!1_{\mathcal{T}}(\{x_{1},...,x_{i}\})u^{\!\{x_{1},...,x_{i}\}\!}d(x_{1},...,x_{i}).
\end{eqnarray*}
Thus, comparing with (\ref{eq:Poissonmeas}), $f$ is indeed a probability
density (with respect to $\mu$) of a Poisson RFSs with intensity
function $u$.

\subsection{Infinite mixtures models with Dirichlet process}

Recent advances in Bayesian nonparametric modeling (BNP) (e.g., see
\cite{Ghosh_bk03_Bayesian,Hjort_etal_bk10_bayesian}) addresses the
unknown number of clusters by introducing a nonparametric prior distribution
on the parameters. One way to motivate the Bayesian nonparametric
setting is to reconsider the mixture likelihood in Eq (\ref{eq:mm_lik-1}).
Let $\pi_{1:K}\sim\dirpdf\left(\cdot|\eta\right),\phi_{k}\iid h,\, k=1,\ldots,K$
where $\dirpdf\left(\cdot|\eta\right)$ is the symmetric Dirichlet
distribution defined before in section \ref{sub:Finite-Bayesian-mixture},
and construct an atomic measure: 
\begin{align}
G & =\sum_{k=1}^{K}\pi_{k}\delta_{\phi_{k}}\label{eq:bmm_stick-1}
\end{align}
 where $\delta_{\phi_{k}}$ denotes the Dirac measure concentrated
at $\phi_{k}$. Note that for a region $A$ on the parameter space,
$G\left(A\right)=\sum_{k=1}^{K}\pi_{k}\bone_{A}(\phi_{k})$. The conditional
distribution for $x$ given $G$ is
\begin{align*}
p\left(x\gv G\right) & =\int f\left(x\gv\phi\right)G\left(d\phi\right)=\int f\left(x\gv\phi\right)\sum_{k=1}^{K}\pi_{k}\delta_{\phi_{k}}\left(d\phi\right)\\
 & =\sum_{k=1}^{K}\pi_{k}\int f\left(x\gv\phi\right)\delta_{\phi_{k}}\left(d\phi\right)=\sum_{k=1}^{K}\pi_{k}f\left(x\gv\phi_{k}\right)
\end{align*}
which identically recovers the likelihood form in Eq (\ref{eq:mm_lik-1}).
Hence, the generative likelihood for the data point $x$ can be equivalently
expressed as: $x\sim f\left(\cdot\gv\phi\right)$ where $\phi\sim G$.
Under this random measure formalism, inference amounts to deriving
the posterior distribution for $G$.

To model an unknown number of clusters, let $\Xi$ be a Poisson RFS
on $\Omega\times\mathbb{R}{}^{+}$, with intensity function $v(\phi,w)=\eta h(\phi)w^{-1}e^{-w}$,
where $\eta>0$, and $h$ is a probability density on $\Omega$. Then
the random measure

\begin{align}
G & =\frac{1}{\bar{w}}\sum_{(\phi,w)\in\Xi}w\delta_{\phi}\label{eq:bmm_stick-1-1}
\end{align}
where $\bar{w}=\sum_{(\phi,w)\in\Xi}w$, is distributed according
to the \emph{Dirichlet process} \cite{Ferguson_73bayesian,Jordan_10hierarchical,Lin_etal_10construction},
i.e.%
\footnote{We note that commonly the Dirichlet process is expressed with a measure
instead of its density, i.e., we could otherwise write $G\sim\dirpdf\left(\eta,H\right)$
where $H$ is a base measure whose density is $h$. However, the use
of the density does not compromise the correctness in this paper,
hence we equivalently use the notation $G\sim\dirpdf\left(\eta,h\right)$
when the density $h$ is the direct object of interest such as the
commonly used likelihood Gaussian in signal processing.%
} $G\sim\dirproc\left(\eta,h\right)$. The RFS $\Xi$ captures the
unknown number of clusters as well as the parameters of the clusters.
This suggests an elegant and tractable%
\footnote{By `tractable' we mean that the posterior is also a Dirichlet process.%
} prior for $G$ is the Dirichlet proces.

Briefly, a Dirichlet process $\dirproc\left(\eta,h\right)$ is a distribution
over random probability measures on the parameter space $\Omega$
and is specified by two parameters: $\eta>0$ is the \emph{concentration}
parameter, and $h$ is the base distribution \cite{Ferguson_73bayesian}.
The terms `Dirichlet' and `base distribution' come from the fact that
for any finite partition of the parameter space $\Omega$, the random
vector obtained by applying $G$ on this partition is distributed
according to a Dirichlet distribution parametrized by $\eta h$. More
concisely, we say $G$ is distributed according to a Dirichlet process,
written as $G\sim\dirproc\left(\eta,h\right)$ if for any \emph{arbitrary}
partition $\left(A_{1},\ldots,A_{m}\right)$ of the space $\Omega$,
$\left(G\left(A_{1}\right),\ldots,G\left(A_{m}\right)\right)\sim\dirpdf\left(\eta\left\langle h,1_{A_{1}}\right\rangle ,\ldots,\eta\left\langle h,1_{A_{m}}\right\rangle \right)$.
The Dirichlet process possesses an extremely attractive conjugate
property, also known as the \emph{Polya urn characterization \cite{Blackwell_MacQueen_73ferguson}}:
let $\phi_{1},\ldots,\phi_{m}$ be i.i.d. samples drawn from $G$,
then 
\begin{align}
 & p\left(\phi_{m}=\phi\gv\phi_{1},\ldots,\phi_{m-1}\right)\label{eq:DP_Polya}\\
 & \qquad\qquad=\frac{\eta h\left(\phi\right)}{m-1+\eta}+\frac{1}{m-1+\eta}\sum_{i=1}^{m-1}1_{\phi_{i}}\left(\phi\right)
\end{align}
Using $G$ as a nonparametric prior distribution, the data generative
process for an infinite mixture models can be summarized as follows:
\begin{align}
G & \sim\dirproc\left(\eta,h\right)\\
\phi_{i} & \sim G\\
x_{i} & \sim f\left(\cdot\gv\phi_{i}\right)
\end{align}
The recent book \cite{Hjort_etal_bk10_bayesian} provides an excellent
account on the theory and applications of the Dirichlet Process.

Alternatively, the nonparametric measure $G$ can be viewed as a limiting
form of the parametric measure $G$ in Eq (\ref{eq:bmm_stick-1})
when $K\goto\infty$ and the weights $\pi_{1:K}$ are drawn from a
symmetric Dirichlet $\dirpdf\left(\frac{\eta}{K},\ldots,\frac{\eta}{K}\right)$
\cite{Teh_etal_06hierarchical}: 
\begin{align}
G & =\sum_{k=1}^{\infty}\pi_{k}\delta_{\phi_{k}}\label{eq:dp_stick-1}
\end{align}
The representation for $G$ in Eq (\ref{eq:dp_stick-1}) is known
as the \emph{stick-breaking} representation, where $\phi_{k}\iid h,k=1,\ldots,\infty$
and $\pi_{1:\infty}$ are the weights constructed through a `stick-breaking'
process \cite{Sethuraman_94constructive}. Imagine we are given a
stick of length $1$, if we infinitely break this stick into small
pieces and assigned each piece to $\pi_{k}$, then clearly, $\sum_{k=1}^{\infty}\pi_{k}=1$.
Since the support of a Beta distribution is between $0$ and $1$,
one may repeatedly sample a value from a Beta distribution and use
this proportion as a principled way to break the stick. Formally,
we construct the infinite dimensional vector $\pi_{1:\infty}$ as
follows:
\begin{align*}
v_{k}\iid & \betapdf\left(1,\eta\right),k=1,\ldots,\infty\\
\pi_{k}= & v_{k}\prod_{s<k}\left(1-v_{s}\right)
\end{align*}
It can be shown that with probability one $\sum_{k=1}^{\infty}\pi_{k}=1$,
and we denote this process as $\pi_{1:\infty}\sim\text{GEM}\left(\eta\right)$
(e.g., see \cite{Hjort_etal_bk10_bayesian,phung_bnp13} for details).

\section{Dirichlet Poisson RFS Mixture Models}

\subsection{Bayesian inference with Poisson RFS\label{sub:Bayesian-inference-with-PoissonRFS}}

In the previous section we see how Poisson-RFS are used to derive
tractable priors, in this section we use Poisson-RFS to develop a
tractable data model. Central to Bayesian analysis is the characterization
of the posterior distribution and the predictive density that expresses
the likelihood of a new data point upon the update of the posterior
distribution.

We start by introducing some necessary notations. Let $f\left(\cdot\gv\Psi\right)$
be a parametric distribution. Occasionally, we use the parameter $\Psi$
to index the distribution $f_{\Psi}$. For example, $f$ is a Gaussian
distribution, then $\Psi=\left(\bmu,\Sigma\right)$ specifies the
mean and covariance matrix. Unless otherwise stated, we further use
$h\left(\cdot\gv\gamma\right)$ to denote the conjugate prior for
$f$ in the sense that the posterior distribution $p\left(\Psi\gv x,\gamma\right)\propto f\left(x_{i}\gv\Psi\right)h\left(\Psi\gv\gamma\right)$
also has the same form as $h$ (with a new parameters $\gamma'$).
For example if $f$ is a Gaussian with unknown mean and fixed variance,
then $h$ is a Gaussian, or if $f$ is Poisson, then $h$ is Gamma
(e.g., see \cite{Gelman_etal_bk03_Bayesian}). 

As described previously, an RFS is a \emph{random point pattern}.
What distinguishes a RFS from a classic random vector-valued random
variable is that the number of points, or elements, is random; and
the points themselves are random and unordered, or simply, an RFS
is a \emph{finite-set-valued random variable} \cite{Vo_phd08_random}.
An RFS $X$ can be fully parametrized by a discrete probability distribution
to specify the cardinality of $X$ and a family of joint distributions
to describe the distribution of values of the points. 

To facilitate our exposition in the sequel \emph{we express a} \emph{Poisson
RFS} $X$\emph{ explicitly as an RFS whose cardinality distribution
follows a Poisson distribution with the rate $\lambda$ and elements
$x$ of $X$ are independently and identically distributed (i.i.d)
according to a probability distribution $f_{\Psi}$ and write $X\sim\poissrfs\left(\lambda,f_{\Psi}\right)$.}

A Poisson RFS $X$ can be sampled as follows: $X=\emptyset,n\sim\poissonpdf\left(\lambda\right)$,
then for $i=1,\ldots,n$ we set $X=X\bigcup\left\{ x_{i}\right\} $
where $x_{i}\iid f_{\Psi}$ and $\poissonpdf\left(\lambda\right)$
is a standard Poisson distribution with mean rate $\lambda$. Assume
unit volume $\mathcal{K}=1$, we express Eq (\ref{eq:Poissondensity})
for Poisson-RFS likelihood density as:
\begin{align*}
p(X\gv\lambda,f_{\Psi})=e^{-\lambda}\lambda^{|X|}f_{\Psi}^{X}
\end{align*}
And when we wish to express the elements of $X$ explicitly as $X=\left\{ x_{1},\ldots,x_{n}\right\} $,
this likelihood density becomes \cite{mahler2007statistical}:
\begin{align}
p\left(X=\left\{ x_{1},\ldots,x_{n}\right\} \gv\lambda,f_{\Psi}\right)=e^{-\lambda}\lambda^{n}\prod_{i=1}^{n}f_{\Psi}\left(x_{i}\right)\label{eq:Poisson-RFS-density}
\end{align}
By convention, when $X$ is an empty set, the RHS reduces to $e^{-\lambda}$.
We note that $X$ is parametrized by two parameters $\lambda$ and
$\Psi$; Let us write them jointly as $\theta=\left(\lambda,\Psi\right)$.
Bayesian inference for Poisson-RFS requires the specification of the
prior distribution over $\theta$. \emph{Furthermore we wish to develop
a} \emph{conjugate prior so that the posterior has the same form as
the prior distribution}. The following proposition summaries our result.
\begin{prop}
Let\emph{ $X\sim\poissrfs\left(\lambda,f_{\Psi}\right)$}, and $h(\cdot\gv\gamma)$
be a conjugate prior of $f_{\Psi}$. Then the distribution given by
\begin{align}
p\left(\lambda,\Psi\gv\alpha,\beta,\gamma\right) & =\frac{\beta^{\alpha}}{\Gamma\left(\alpha\right)}\lambda^{\alpha-1}e^{-\beta\lambda}h\left(\Psi\gv\gamma\right)\label{eq:Bayesian_RFS-conjugate}
\end{align}
is the conjugate prior for $X$, in the sense that the posterior distribution
$p\left(\lambda,\Psi\gv X,\alpha,\beta,\gamma\right)$ has the same
form as (\ref{eq:Bayesian_RFS-conjugate}).\end{prop}
\begin{IEEEproof}
To prove this, we note that the Gamma distribution is a conjugate
prior for a Poisson distribution and $H$ is conjugate to $F$, hence
our first guess is that this conjugate structure will carry on for
a Poisson-RFS. And, it turns out that this intuition is indeed correct
as described below. 

To see why, let $\lambda\sim\gammapdf\left(\alpha,\beta\right)$ so
that $p\left(\lambda\gv\alpha,\beta\right)\propto\lambda^{\alpha-1}e^{-\beta\lambda}$
and using Baye's rule, the \textbf{\emph{posterior}} distribution
takes the form
\begin{eqnarray*}
p\left(\lambda,\Psi\gv X,\alpha,\beta,\gamma\right)\propto p\left(X\gv\lambda,\Psi\right)p\left(\lambda,\Psi\gv\alpha,\beta,\gamma\right)\\
\propto e^{-\lambda}\lambda^{|X|}f_{\Psi}^{X}\lambda^{\alpha-1}e^{-\beta\lambda}h\left(\Psi\gv\gamma\right)\\
\propto\left[\lambda^{|X|+\alpha-1}e^{-(\beta+1)\lambda}\right]\left[f_{\Psi}^{X}h\left(\Psi\gv\gamma\right)\right]
\end{eqnarray*}
It is clear that this has the same form as the prior distribution
in Eq (\ref{eq:Bayesian_RFS-conjugate}) since the last term will
results in $h$-like distribution due to conjugacy of $h$ and $f_{\Psi}$.
Given an observed $X$, the rate $\lambda$ now follows $\gammapdf\left(|X|+\alpha,\beta+1\right)$
and $\Psi$ follows $h(\cdot\gv\gamma')$ where $\gamma'$ is the
posterior parameter resulting from $f_{\Psi}^{X}h\left(\Psi\gv\gamma\right)$
due the conjugacy of $h$ and $f_{\Psi}$ and have values depending
on specification of $h$ and $f_{\Psi}$. 
\end{IEEEproof}
By induction, the posterior distribution after observing $N$ set-valued
observation $\left\{ X_{1},\ldots,X_{N}\right\} $ is
\begin{align*}
p\left(\lambda,\Psi\gv X_{1},\ldots,X_{n},\alpha,\beta,\gamma\right)\,\,\,\,\qquad\qquad\qquad\qquad\\
\propto\left[\lambda^{\sum_{i=1}^{N}|X_{i}|+\alpha-1}e^{-(\beta+N)\lambda}\right]\left[\prod_{i=1}^{N}f_{\Psi}^{X_{i}}h\left(\Psi\gv\gamma\right)\right]
\end{align*}
The posterior for $\lambda$ is now $\gammapdf\left(\alpha_{N},\beta_{N}\right)$
with $\alpha_{N}=\alpha+\sum_{i=1}^{N}|X_{i}|,\beta_{N}=\beta+N$;
whereas $\Psi$ follows $h(\cdot\gv\gamma_{N})$ where $\gamma_{N}$
is posterior parameter obtained from evaluating $\prod_{i=1}^{N}f_{\Psi}^{X_{i}}h\left(\Psi\gv\gamma\right)$.

As in a standard Bayesian analysis problem, given the observed data
$D=\left\{ X_{1},\ldots,X_{N}\right\} $ it is important to be able
to specify the \emph{predictive} likelihood of an unseen observation
$X$ for a prediction task. For our mixture model developed in sequel,
we use this likelihood in the Gibbs sampler to assess the likelihood
of data points being assigned to cluster components. It turns out
that this predictive density is also tractable for our Bayesian Poisson-RFS
case. With a small effort of manipulation, this can be shown to be:

\begin{align}
 & p\left(X\gv X_{1:N},\alpha,\beta,\gamma\right)\nonumber \\
 & \,=\int\int p\left(X\gv\lambda,\Psi\right)p\left(\lambda,\Psi\gv X_{1:N}\right)d\lambda d\Psi\nonumber \\
 & \,=\left[\frac{\Gamma(\alpha_{N}+|X|)\beta_{N}^{\alpha_{N}}}{\Gamma(\alpha_{N})\left(\beta_{N}+1\right)^{\alpha_N + |X|}}\right]\left[\int f_{\Psi}^{X}h\left(\Psi\gv\gamma_{N}\right)d\Psi\right]\label{eq:predictive}
\end{align}
Again, depending on the specific forms for $h$ and $f_{\Psi}$, the
last term can be evaluated analytically (see \cite{bernardo2009bayesian}
for several examples).

\subsection{The Dirichlet Poisson RFS mixture model}

The intuition for our proposed Dirichlet Poisson RFS Mixture Model
(DP-RFS) is that \emph{each mixture component is now a Poisson-RFS},
hence the model's support is now the space of finite sets. Therefore,
we model set-valued data as random quantities and estimate a mixture
density with an infinite number mixture components over these data.
Since the data likelihood is a mixture of Poisson RFS densities, each
mixture component is parameterised by the tuple $\phi_{k}=\left(\lambda_{k},\Psi_{k}\right)$.
To do so, let $G$ follows a Dirichlet process whose base distribution
is a conjugate prior specified in Eq (\ref{eq:Bayesian_RFS-conjugate}).
Using $G$ as a nonparametric prior distribution, the data generative
process for our model for $N$ set-valued observations $\left\{ X_{1},\ldots,X_{N}\right\} $
can be summarized as follows: 
\begin{align}
G & \sim\dirproc\left(\eta,h'\right)\\
\phi_{i}=\left(\lambda_{i},\Psi_{i}\right) & \sim G,\\
X_{i} & \sim\poissrfs\left(\lambda_{i},f_{\Psi_{i}}\right)
\end{align}
where 
\begin{align*}
h'\left(\lambda,\Psi\gv\alpha,\beta,\gamma\right)=\frac{\beta^{\alpha}}{\Gamma\left(\alpha\right)}\lambda^{\alpha-1}e^{-\beta\lambda}h\left(\Psi\gv\gamma\right)
\end{align*}
taken as the conjugate prior developed in Eq (\ref{eq:Bayesian_RFS-conjugate}).
Our Dirichlet Poisson RFS mixture model then specifies an infinite
mixture over a set-valued observation $X$ as:
\begin{align}
p\left(X\gv\pi_{1:\infty},\phi_{1:\infty}\right) & =\sum_{k=1}^{\infty}\pi_{k}\left[e^{-\lambda_{k}}\lambda_{k}^{|X|}f_{\Psi_{k}}^{X}\right]\label{eq:}
\end{align}

\subsection{Markov Chain Monte Carlo Inference }

Given only the data $\left\{ X_{1},X_{2},\ldots,X_{N}\right\} $,
the concentration parameter $\eta$ and the parameters $\alpha,\beta,\gamma$
for the base distribution $h$', our task is to infer a posterior
distribution for $\pi_{1:\infty}$ and $\phi_{1:\infty}$. This is
an intractable Bayesian inference problem and an MCMC inference scheme
is needed. A full Gibbs inference similar to the scheme described
in section \ref{sub:Finite-Bayesian-mixture} (cf. Eq \ref{eq:full_Gibbs1}--\ref{eq:full_Gibbs3})
can be developed. For faster convergence, we describe in this section
a collapsed Gibbs inference. We introduce the latent cluster indicators
$z_{i}$ to explicitly indicate the mixture component to which the
data point $X_{i}$ being assigned to and sample them directly, whereas
$\pi_{1:\infty}$ and $\phi_{1:\infty}$ will be integrated out.

Using the stick-breaking represention for the Dirichlet process the
data generative process can be now equivalently expressed as:

\noindent%
\fbox{\begin{minipage}[t]{0.98\columnwidth}%
\begin{align*}
\left(\lambda_{k},\Psi_{k}\right) & \iid h'\left(\cdot\gv\alpha,\beta,\gamma\right)\text{ for }k=1,2,\ldots\\
\pi_{k} & =v_{k}\prod_{s<k}\left(1-v_{s}\right)\text{ where }v_{k}\iid\betapdf\left(1,\eta\right)\\
\text{For } & i=1,\ldots,N\\
 & z_{i}\sim\text{Discrete}\left(\pi_{1:\infty}\right)\\
 & X_{i}\sim\poissrfs\left(\lambda_{z_{i}},\Psi_{z_{i}}\right)\\
\text{End}
\end{align*}
where the extra notation $\text{Discrete}\left(\cdot\right)$ denote
a discrete distribution whose support is the set of positive integers.%
\end{minipage}}

\medskip{}
Our aim is to perform posterior inference on the $p\left(z_{1:}{}_{N}\gv X_{1:N},\Phi\right)$,
where $\Phi=\left\{ \alpha,\beta,\gamma,\eta\right\} $ is the set
of so-called \emph{hyper-parameters}. 

This inference can be carried out under a Gibbs sampling scheme using
the Polya urn characterization of the Dirichlet process \cite{Blackwell_MacQueen_73ferguson},
otherwise also known as the Chinese restaurant process \cite{Pitman_bk06_combinatorial}.
The structure of our inference scheme follows the work \cite{Neal_00markov}
for generic Gibbs inference for Dirichlet Process Mixture model. Central
to this Gibbs inference scheme is the conditional distribution $p\left(z_{i}\gv z_{-i},X_{1:N},\Phi\right)$
from which one iteratively scans through each $z_{i}$ and sample
it. This conditional distribution can be expressed as follows using
Bayes' rule and recall that the notation $z_{-i}$ denotes the set
of all assignment indicators except $z_{i}$, and likewise for $X_{-i}$:
\begin{align}
 & p\left(z_{i}=k\gv z_{-i},X_{1:N},\Phi\right)\nonumber \\
 & =p\left(z_{i}=k\gv z_{-i},X_{i},X_{-i},\Phi\right)\nonumber \\
 & \propto p\left(X_{i}\gv z_{i}=k,z_{-i},X_{-i},\Phi\right)p\left(z_{i}=k\gv z_{-i},X_{-i},\Phi\right)\nonumber \\
 & \propto p\left(X_{i}\gv z_{i}=k,z_{-i},X_{-i},\Phi\right)p\left(z_{i}=k\gv z_{-i},\Phi\right)\label{eq:Gibbs_equation}
\end{align}
Note that in the last term $X_{-i}$ has been removed due to the fact
that $z_{i}$ is conditionally independent of $X_{-i}$ given $z_{-i}$
in the absence of $X_{i}$. Due to the Polya urn characterization
of the Dirichlet process as described in Eq (\ref{eq:DP_Polya}) the
second term $p\left(z_{i}=k\gv z_{-i},\Phi\right)$ can be written
as:
\begin{align*}
p\left(z_{i}=k\gv z_{-i},\Phi\right) & =\begin{cases}
\frac{n_{-i,k}}{n-1+\eta} & \text{ if }k\text{ exists}\\
\frac{\eta}{n-1+\eta} & \text{ if }k\text{ is new}
\end{cases}
\end{align*}
where we recall that ${\textstyle n_{-i,k}=\sum_{j=1,j\neq i}^{N}}1_{z_{j}}(k)$.
This is also known the \emph{Chinese Restaurant Process} in combinatorial
stochastic process \cite{Pitman_bk06_combinatorial}. This expression
illustrates the \emph{clustering property} induced by the mode: a
future data observation is more likely to return to an existing cluster
with a probability proportional to its popularity $n_{-i,k}$, but
it is also flexible enough to pick on a new value if needed as data
grows beyond the complexity that current model can explain. Furthermore,
the number of clusters grow at $O\left(n\log\gamma\right)$ under
the Dirichlet process prior \cite{Ferguson_73bayesian,Antoniak_74mixtures}.

The first term $p\left(X_{i}\gv z_{i}=k,z_{-i},X_{-i},\Phi\right)$
in Eq (\ref{eq:Gibbs_equation}) can be recognized as a form of predictive
likelihood with respect to the mixture component $k$, where the predictive
likelihood for unseen data point under Bayesian inference for Poisson
RFS has been developed previously in section \ref{sub:Bayesian-inference-with-PoissonRFS}
(cf. Eq \ref{eq:predictive}) 

$p\left(X_{i}\gv z_{i}=k,z_{-i},X_{-i},\Phi\right)=$ {\small{}
\begin{align*}
\int\int p\left(X_{i}\gv\lambda_{k},\Psi_{k}\right)p\left(\lambda_{k},\Psi_{k}\gv\left\{ X_{j}:z_{j}=k,j\neq i\right\} \right)d\lambda_{k}d\Psi_{k}
\end{align*}
}and we shall denote this likelihood as $f_{k}\left(X_{i};X_{-i}\right)$.
Gibbs sampling then simply involves iteratively sampling $z_{1},\ldots,z_{N}$
as summarized in Algorithm \ref{alg:Collapsed-Gibbs-inference}. 

\begin{algorithm}
\textbf{Input}
\begin{itemize}
\item Set-valued observations $X_{1},\ldots,X_{N}$
\item Concentration parameter $\eta$ and prior parameters $\alpha,\beta,\gamma$
\item Number of Gibbs samples $L$.
\end{itemize}
\textbf{Collapse Gibbs inference}
\begin{enumerate}
\item Initialize a random number of mixtures $K$ (say $1$)
\item Initialize randomly $z_{1}^{\left(0\right)},\ldots,z_{N}^{\left(0\right)}$
so that $1\leq z_{i}^{\left(0\right)}\leq K$
\item For $l=1$ to $L$

~~~~~For $i=1$ to $n$ sample $z_{i}$ from

~~~~~~~~~~$p(z_{i}^{\left(l\right)}=k\mid z_{-i}^{\left(l-1\right)},X_{1:N},\gamma)$

~~~~~~~~~~~~~~~$\propto\begin{cases}
n_{-i,k}f_{k}\left(X_{i};X_{-i}\right) & \text{ if }k\leq K\\
\eta f\left(X_{i}\right) & \text{ if }k=K+1
\end{cases}$

~~~~~~~~~~If $z_{i}^{\left(l\right)}=K+1$, set $K\leftarrow K+1$

\item Remove any empty mixture component and decrease $K$ accordingly.
\end{enumerate}
\textbf{Output}:
\begin{itemize}
\item The number of mixture components learned $K$.
\item $L$ Gibbs samples $\left\{ z_{1}^{\left(l\right)},\ldots,z_{N}^{\left(l\right)}\right\} _{l=1}^{L}$
for the cluster indicators.
\end{itemize}
\protect\caption{Collapsed Gibbs inference for the proposed Dirichlet Poisson RFS Mixture
Models.\label{alg:Collapsed-Gibbs-inference}}
\end{algorithm}

Note in this algorithm that when $z_{i}$ takes on a new cluster,
i.e., $z_{i}=K+1$ the predictive likelihood $f\left(X_{i}\right)$
is simply an integration over the prior distribution without observation
any data point in this newly mixture component yet, i.e., 
\begin{align*}
f\left(X_{i}\right) & =\int\int p\left(X_{i}\gv\lambda,\Psi\right)p\left(\lambda,\Psi\gv\alpha,\beta,\gamma\right)d\lambda d\Psi
\end{align*}
In practice, we discard some initial Gibbs samples, a strategy commonly
known as burn-in period in MCMC literature. In our experiment, to
provide robustness we also sample the concentration parameter $\eta$
according the procedure described in \cite{Escobar_West_95bayesian};
however it is not essential to understanding the Gibbs inference routine
here, hence its description will be skipped.

\section{Numerical Results}

\begin{figure*}[t]
\begin{centering}
\includegraphics[width=0.7\textwidth,height=0.25\paperheight]{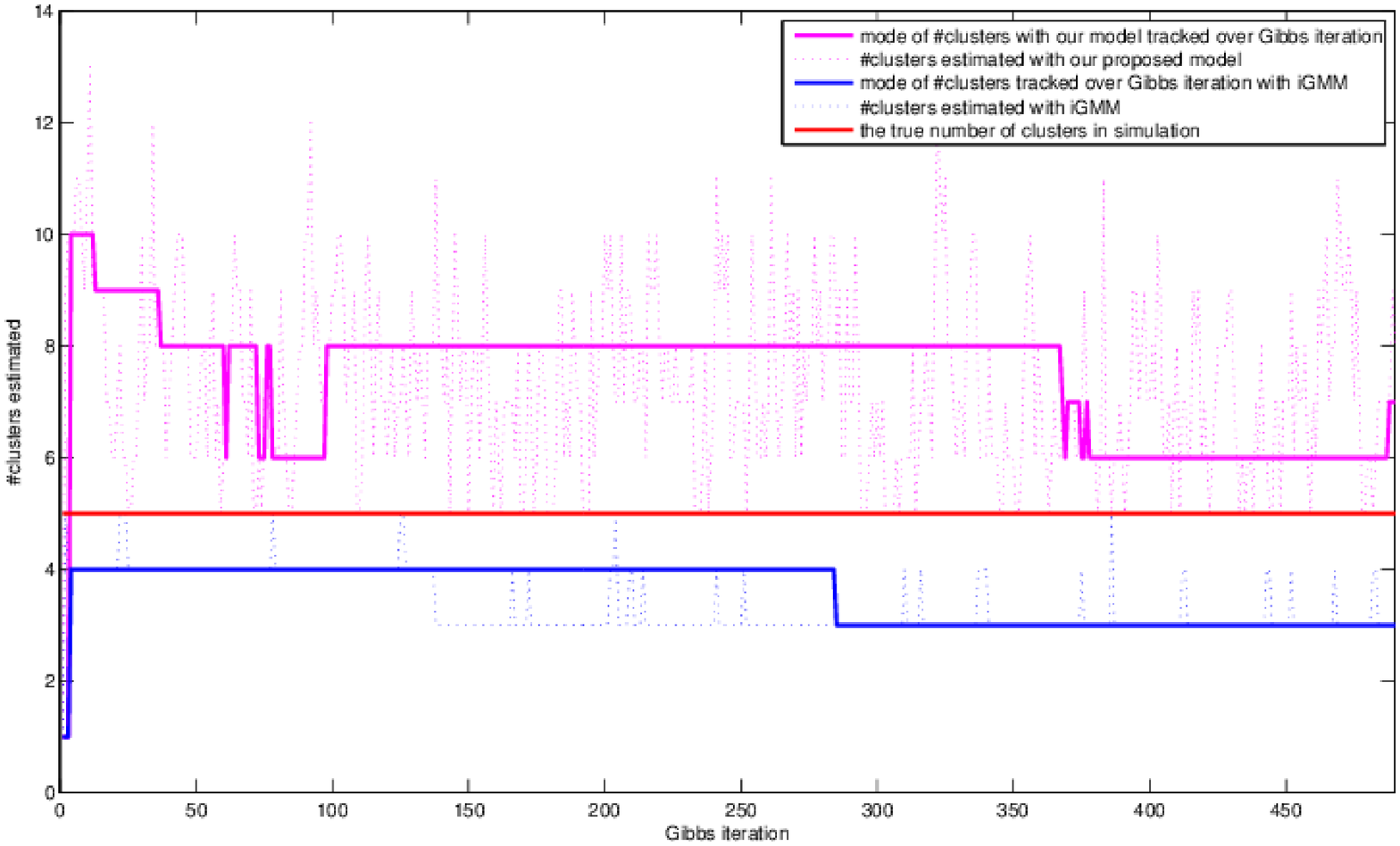}
\par\end{centering}

\begin{centering}
\includegraphics[width=0.3\paperwidth]{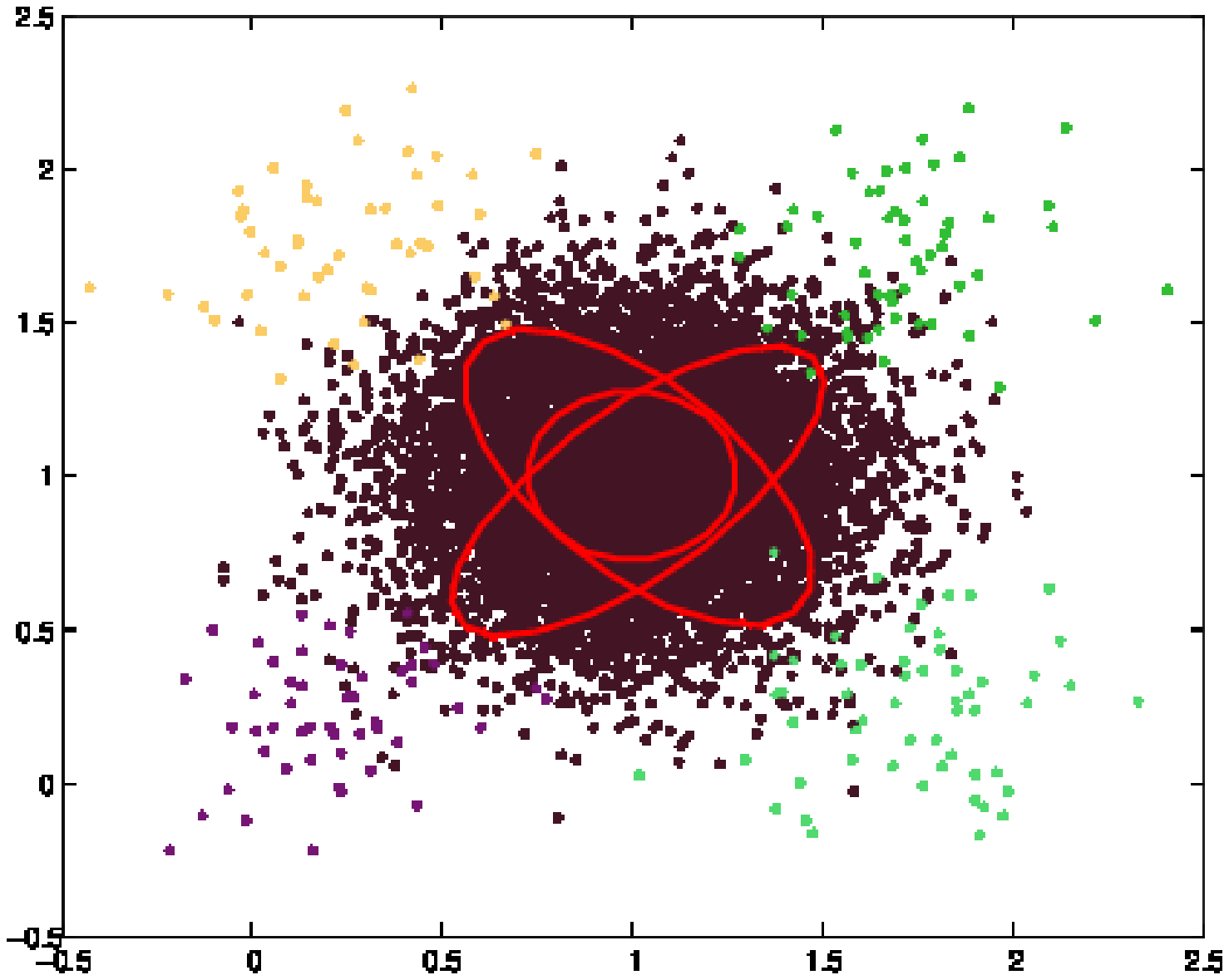}~~~~~~~\includegraphics[width=0.3\paperwidth]{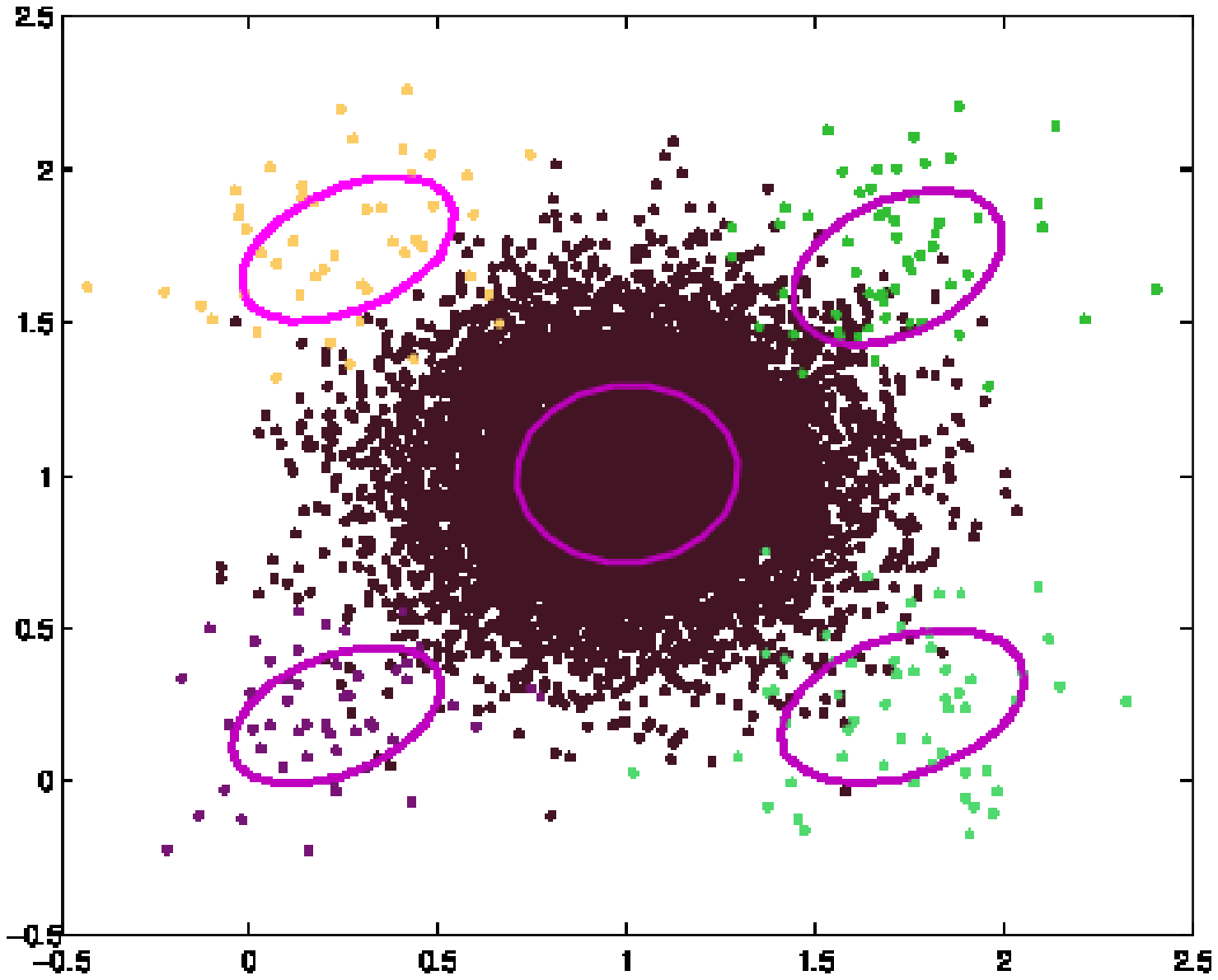}
\par\end{centering}

\protect\caption{Numerical results. Five clusters created with one dominant clusters
at the center (see main text for more details). Top: the estimated
number of clusters varies with Gibbs iteration. Bottom-left: result
of the current state-of-the-art infinite mixture of Gaussian \cite{rasmussen1999infinite}
which misses four outlier clusters completely; Bottom-right: results
from our DP-RFS mixture which correctly identifies all five clusters.
\label{fig:simulation_result}}
\end{figure*}

This secion demonstrates the key properties of the proposed model
via two numerical studies.  We focus on one typical phenomenon in
data modelling known as data clustering with extremely unbalanced
datasets -- an open challenging problem in data clustering analysis
\cite{he2009learning,yen2009cluster}. We construct \emph{five} Gaussians
arranged in a star-shape: sitting at the center is a large-variance
Gaussian specified with Poisson rate of $100,$ which \emph{dominates}
the generation of data; four other Gaussians scattered over the four
corners and are specified with an extremely low Poisson rate of $0.5$.
Hence, as seen in Figure \ref{fig:simulation_result}, the data looks
as if it is generated solely by the dominant Gaussian and consequently
this scenario presents a very challenging case to model the other
four `outlier' clusters. This is also known as an imbalanced data
problem in related field of unsupervised learning and data mining
and is frequently encountered in novelty and abnormality detection
problem \cite{he2009learning,yen2009cluster,bishop2006pattern,han2006data}. 

Our baseline comparison is the state-of-the-art infinite Gaussian
mixture model (iGMM) \cite{rasmussen1999infinite} which is a Bayesian
nonparametric version of the classic Gaussian mixture models. This
model can also bypass the model selection problem to automatically
discover the number of clusters from the data. Input to iGMM is vector-valued
data, hence we take the union of set-valued observations as the data
for iGMM. We ensure that the initializations for our DP-RFS model
and iGMM are as similar as possible and ran 500 Gibbs iterations after
a small burn-in period. We keep track of the mode of the number of
clusters as we progress and use the last result as our estimated result
(equivalent to a MAP estimation with Gibbs sample). 

Figure \ref{fig:simulation_result} presents the results of the simulation.
The top figure shows the estimated number of clusters $K$ varies
with Gibbs iteration. We initialize $K=1$ for both iGMM and our model.
Note that iGMM tends to under estimate the number of clusters due
to dominant cluster; our DP-RFS model, on the other hand, tends to
over estimate the number of clusters at first, but gradually approaches
the true number of cluster. This is partially explained by the use
of Poisson RFS likelihood in the model, which provides the flexibility
in creating spurious and skewed clusters to explain the data. 

At termination, iGMM yields three clusters as seen in the bottom-left
of the figure; and completely missed the four outlier clusters. The
two Gaussians with diagonal direction appears to be affected and confused
by the outlier clusters. Our DP-RFS model discovers 6 clusters, however
one has an infinite variance and hence eliminated leaving five clusters
plotted in the bottom-right of Figure \ref{fig:simulation_result}.
Our proposed technique has correctly identified the dominant cluster
and all other four outlier clusters. Further, it estimates the Poisson
rate for the dominant cluster to be $77.32$ and the other four are
$0.34,0.35,0.38$ and $0.37$, which are quite close to the groundtruth.

\begin{figure*}[t]
\begin{centering}
\includegraphics[width=0.28\paperwidth]{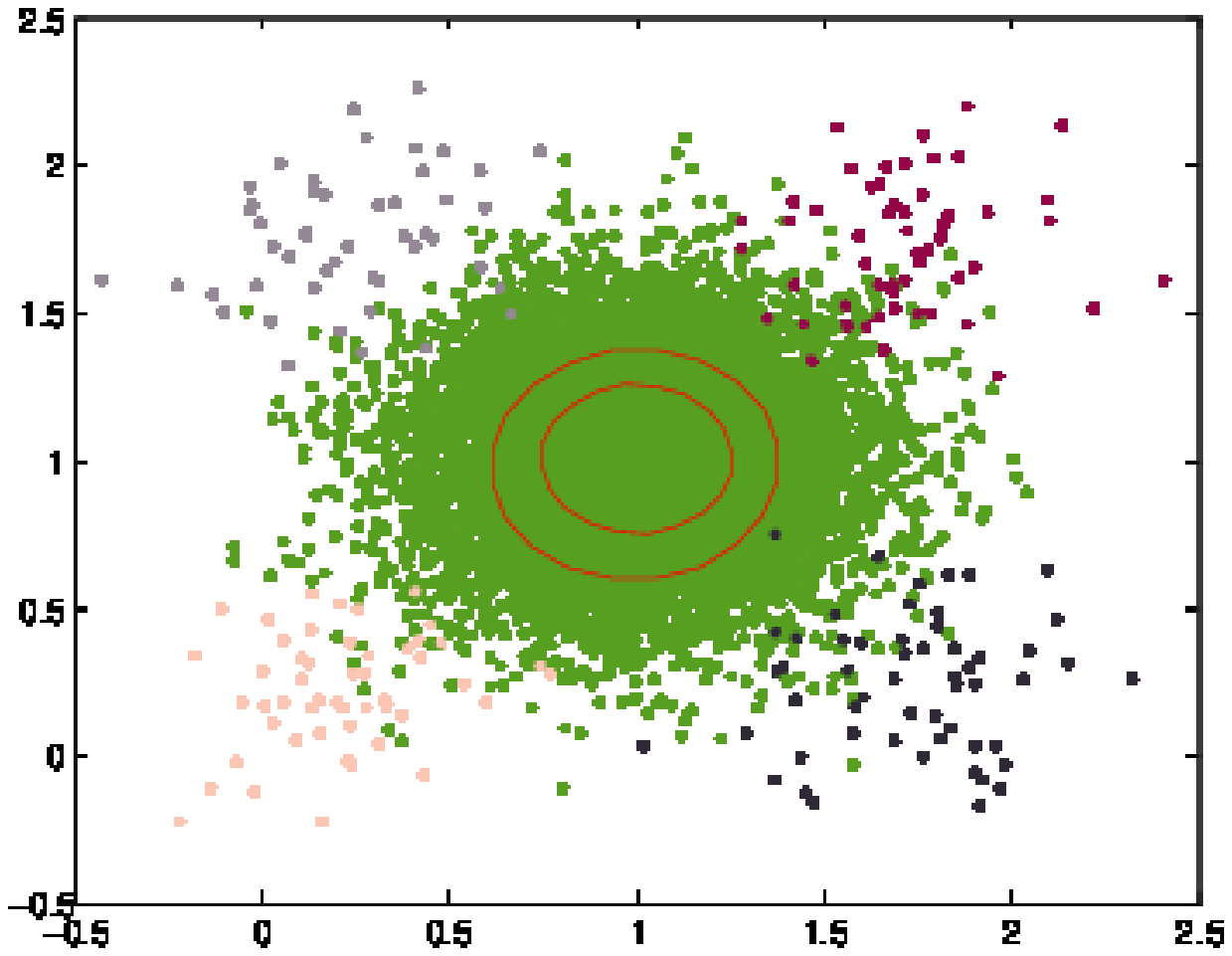}\includegraphics[width=0.28\paperwidth]{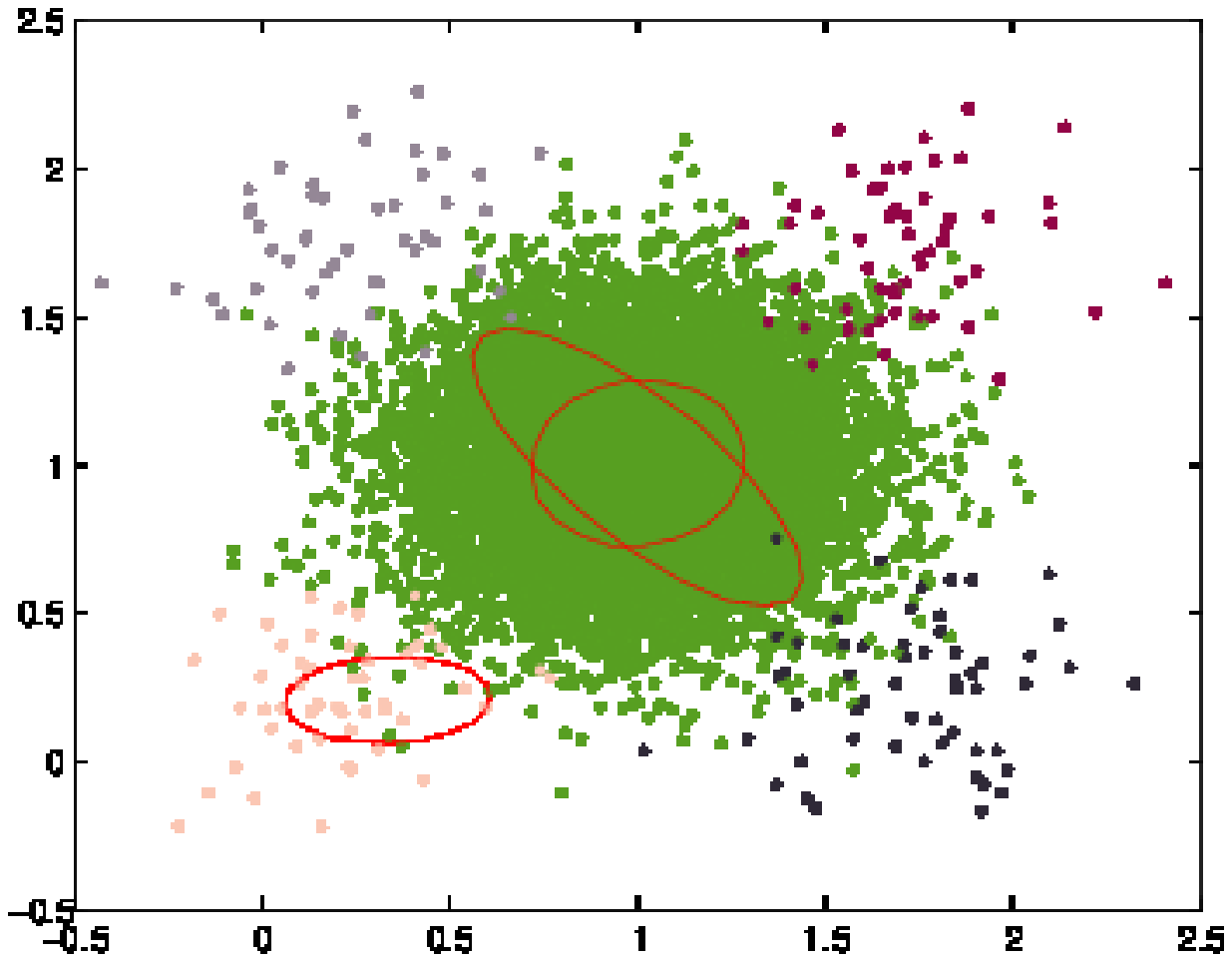}\\
\includegraphics[width=0.28\paperwidth]{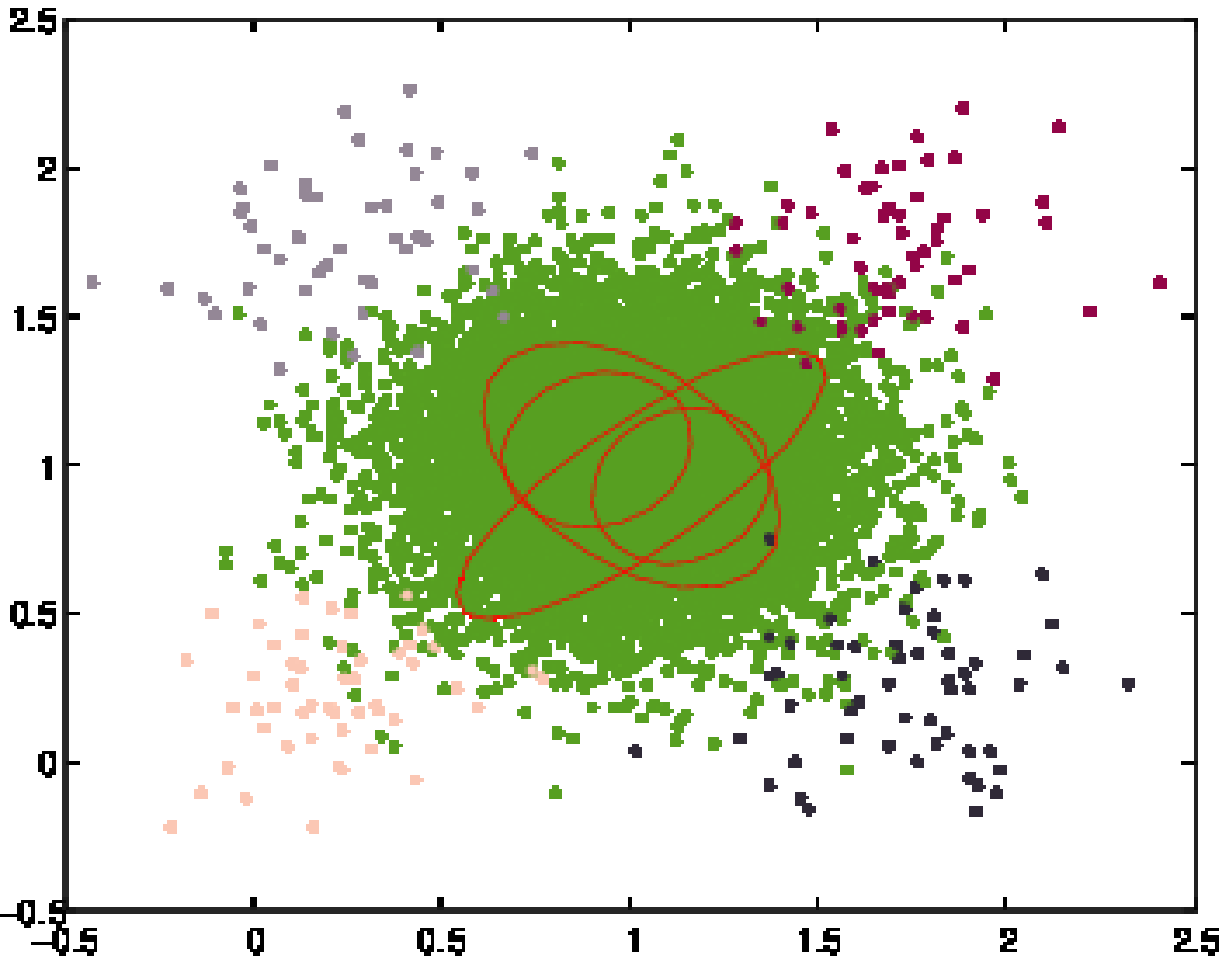}\includegraphics[width=0.28\paperwidth]{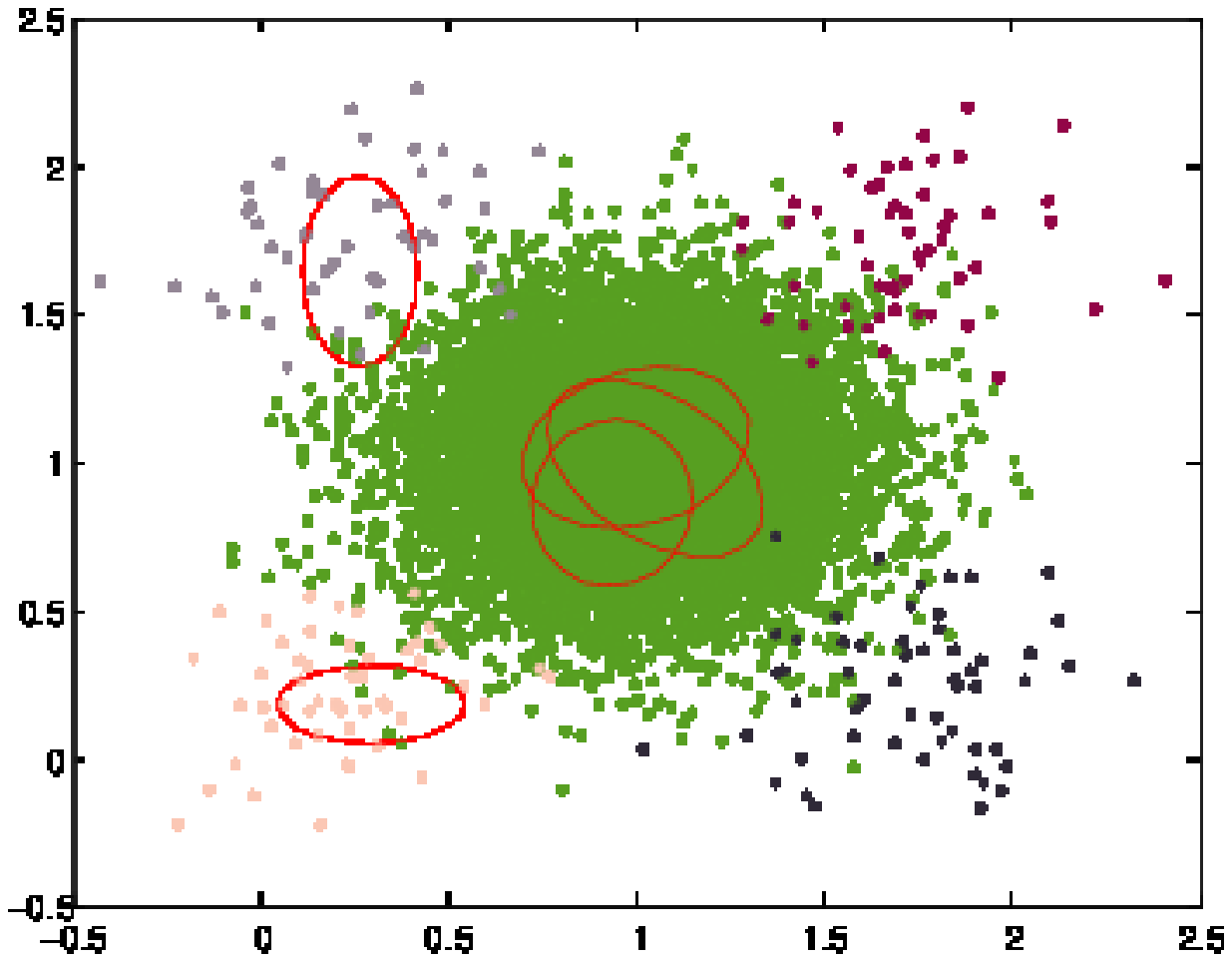}~\includegraphics[width=0.28\paperwidth]{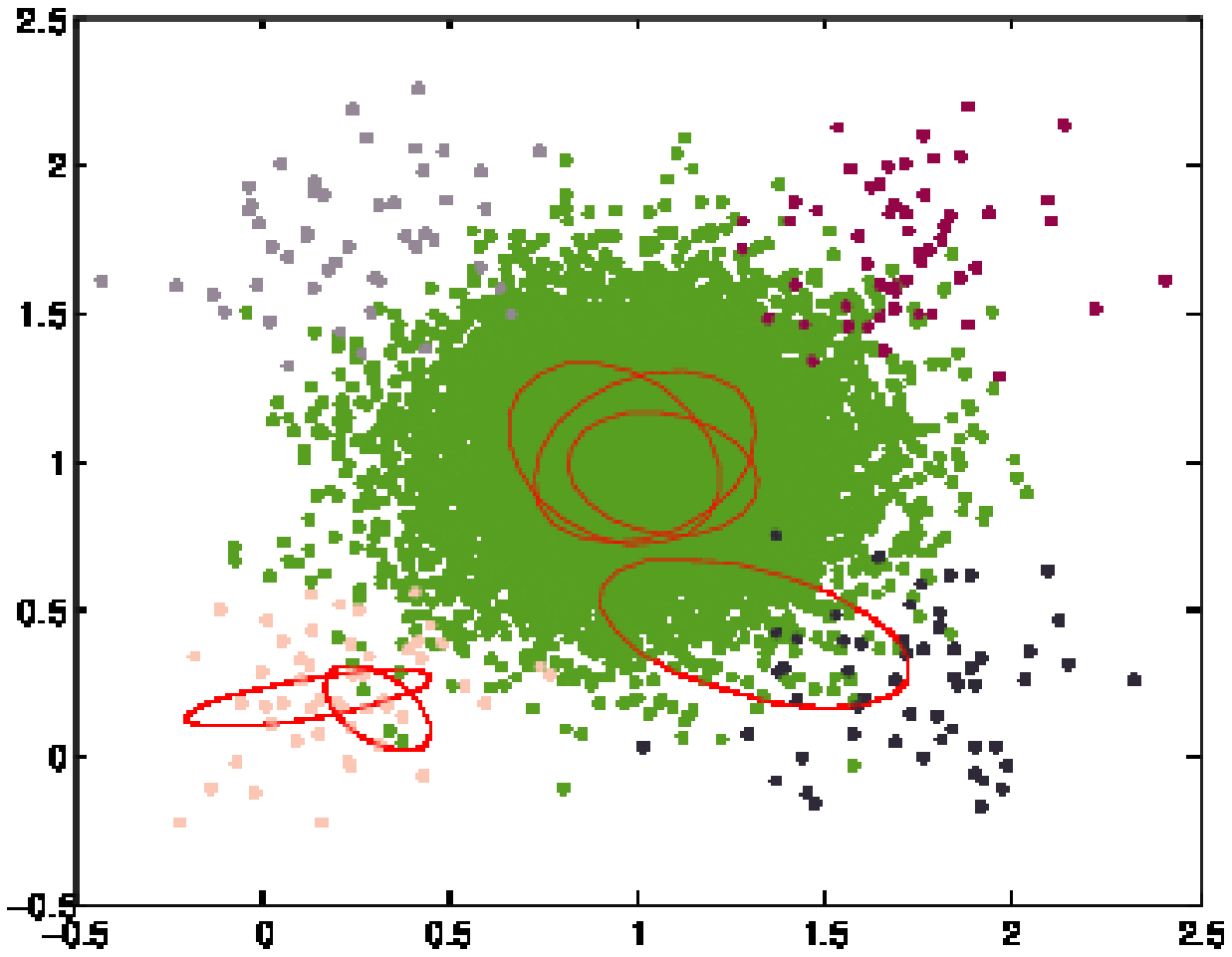}
\par\end{centering}

\protect\caption{Clustering results using finite Mixture of Gaussian model (MoG). In
this case, we need to specify the number of clusters in advances.
From left to right and top to bottom, we set the number of clusters
to be $2,3,4,5,6$ respectively.\label{fig:simulation_result-1}}
\end{figure*}

To illustrate further clustering behaviors in the existence of imbalanced
clusters, we present the results that used the common Mixture of Gaussians
(MoG) for clustering tasks. While iGMM \cite{rasmussen1999infinite}
and our proposed DP-RFS mixture model can automatically infer the
number of clusters $K$ from data, MoG requires us to specify this
number in advance. Figure \ref{fig:simulation_result-1} presents
the results for $K=2,3,4,5$ and $6$. Again, in addition to the fact
that MoG is unable to infer the number of clusters, it suffers a similar
effect as observed in iGMM wherein the existence of the dominant cluster
makes it almost impossible to learn the other four outlier clusters.

\section{Discussion and conclusion}

In this paper we have shown how Poisson RFS can be used to develop
infinite mixture model data clustering. In particular, we developed
a conjugate prior for a Poisson-RFS likelihood with all of the properties
of a typical Bayesian conjugate setting, including its conjugate posterior
distribution and predictive density. Using this result, we constructed
an infinite mixture of Poisson-RFS using the recently developed Dirichlet
process theory for Bayesian nonparametric mixture models. This results
in a new class of statistical models to both signal processing and
machine learning: \emph{it is an infinite mixture over set-valued
data observations} and we term this model the Dirichlet Poisson Random
Finite Set mixture model (DP-RFS). As set-valued observations arises
naturally in everyday analysis tasks, we anticipate that this line
of modelling will accommodate a wide range of applications. The numerical
study presented in this paper has demonstrated the capacity of the
proposed DP-RFS model to tackle the open challenge of modelling and
clustering imbalanced data. Lastly, beyond Poisson-RFS, our framework
opens the door to more general RFS models for data clustering. 

\bibliographystyle{abbrv}
\bibliography{dpresearch,new}

\end{document}

%% file: macros.tex
\global\long\def\sidenote#1{\marginpar{\small\emph{{\color{Medium}#1}}}}

\global\long\def\se{\hat{\text{se}}}

\global\long\def\interior{\text{int}}

\global\long\def\boundary{\text{bd}}

\global\long\def\ML{\textsf{ML}}

\global\long\def\GML{\mathsf{GML}}

\global\long\def\HMM{\mathsf{HMM}}

\global\long\def\support{\text{supp}}

\global\long\def\new{\text{*}}

\global\long\def\stir{\text{Stirl}}

\global\long\def\mA{\mathcal{A}}

\global\long\def\mB{\mathcal{B}}

\global\long\def\mF{\mathcal{F}}

\global\long\def\mK{\mathcal{K}}

\global\long\def\mH{\mathcal{H}}

\global\long\def\mX{\mathcal{X}}

\global\long\def\mZ{\mathcal{Z}}

\global\long\def\mS{\mathcal{S}}

\global\long\def\Ical{\mathcal{I}}

\global\long\def\mT{\mathcal{T}}

\global\long\def\Pcal{\mathcal{P}}

\global\long\def\dist{d}

\global\long\def\HX{\entro\left(X\right)}
 \global\long\def\entropyX{\HX}

\global\long\def\HY{\entro\left(Y\right)}
 \global\long\def\entropyY{\HY}

\global\long\def\HXY{\entro\left(X,Y\right)}
 \global\long\def\entropyXY{\HXY}

\global\long\def\mutualXY{\mutual\left(X;Y\right)}
 \global\long\def\mutinfoXY{\mutualXY}

\global\long\def\given{\mid}

\global\long\def\gv{\given}

\global\long\def\goto{\rightarrow}

\global\long\def\asgoto{\stackrel{a.s.}{\longrightarrow}}

\global\long\def\pgoto{\stackrel{p}{\longrightarrow}}

\global\long\def\dgoto{\stackrel{d}{\longrightarrow}}

\global\long\def\lik{\mathcal{L}}

\global\long\def\logll{\mathit{l}}

\global\long\def\vectorize#1{\mathbf{#1}}

\global\long\def\vt#1{\mathbf{#1}}

\global\long\def\gvt#1{\boldsymbol{#1}}

\global\long\def\idp{\ \bot\negthickspace\negthickspace\bot\ }
 \global\long\def\cdp{\idp}

\global\long\def\das{\triangleq}

\global\long\def\id{\mathbb{I}}

\global\long\def\idarg#1#2{\id\left\{  #1,#2\right\}  }

\global\long\def\iid{\stackrel{\text{iid}}{\sim}}

\global\long\def\bzero{\vt 0}

\global\long\def\bone{\mathbf{1}}

\global\long\def\boldm{\boldsymbol{m}}

\global\long\def\bff{\vt f}

\global\long\def\bx{\boldsymbol{x}}

\global\long\def\bl{\boldsymbol{l}}

\global\long\def\bu{\boldsymbol{u}}

\global\long\def\bo{\boldsymbol{o}}

\global\long\def\bh{\boldsymbol{h}}

\global\long\def\bs{\boldsymbol{s}}

\global\long\def\bz{\boldsymbol{z}}

\global\long\def\xnew{y}

\global\long\def\bxnew{\boldsymbol{y}}

\global\long\def\bX{\boldsymbol{X}}

\global\long\def\tbx{\tilde{\bx}}

\global\long\def\by{\boldsymbol{y}}

\global\long\def\bY{\boldsymbol{Y}}

\global\long\def\bZ{\boldsymbol{Z}}

\global\long\def\bU{\boldsymbol{U}}

\global\long\def\bv{\boldsymbol{v}}

\global\long\def\bn{\boldsymbol{n}}

\global\long\def\bV{\boldsymbol{V}}

\global\long\def\bI{\boldsymbol{I}}

\global\long\def\bw{\vt w}

\global\long\def\balpha{\gvt{\alpha}}

\global\long\def\bbeta{\gvt{\beta}}

\global\long\def\bmu{\gvt{\mu}}

\global\long\def\btheta{\boldsymbol{\theta}}

\global\long\def\blambda{\boldsymbol{\lambda}}

\global\long\def\bgamma{\boldsymbol{\gamma}}

\global\long\def\bpsi{\boldsymbol{\psi}}

\global\long\def\bphi{\boldsymbol{\phi}}

\global\long\def\bpi{\boldsymbol{\pi}}

\global\long\def\bomega{\boldsymbol{\omega}}

\global\long\def\bepsilon{\boldsymbol{\epsilon}}

\global\long\def\btau{\boldsymbol{\tau}}

\global\long\def\bvarphi{\boldsymbol{\varphi}}

\global\long\def\realset{\mathbb{R}}

\global\long\def\realn{\realset^{n}}

\global\long\def\integerset{\mathbb{Z}}

\global\long\def\natset{\integerset}

\global\long\def\integer{\integerset}

\global\long\def\natn{\natset^{n}}

\global\long\def\rational{\mathbb{Q}}

\global\long\def\rationaln{\rational^{n}}

\global\long\def\complexset{\mathbb{C}}

\global\long\def\comp{\complexset}

\global\long\def\compl#1{#1^{\text{c}}}

\global\long\def\and{\cap}

\global\long\def\compn{\comp^{n}}

\global\long\def\comb#1#2{\left({#1\atop #2}\right) }

\global\long\def\nchoosek#1#2{\left({#1\atop #2}\right)}

\global\long\def\param{\vt w}

\global\long\def\Param{\Theta}

\global\long\def\meanparam{\gvt{\mu}}

\global\long\def\Meanparam{\mathcal{M}}

\global\long\def\meanmap{\mathbf{m}}

\global\long\def\logpart{A}

\global\long\def\simplex{\Delta}

\global\long\def\simplexn{\simplex^{n}}

\global\long\def\dirproc{\text{DP}}

\global\long\def\ggproc{\text{GG}}

\global\long\def\DP{\text{DP}}

\global\long\def\ndp{\text{nDP}}

\global\long\def\hdp{\text{HDP}}

\global\long\def\gempdf{\text{GEM}}

\global\long\def\rfs{\text{RFS}}

\global\long\def\bernrfs{\text{BernoulliRFS}}

\global\long\def\poissrfs{\text{PoissonRFS}}

\global\long\def\grad{\gradient}
 \global\long\def\gradient{\nabla}

\global\long\def\partdev#1#2{\partialdev{#1}{#2}}
 \global\long\def\partialdev#1#2{\frac{\partial#1}{\partial#2}}

\global\long\def\partddev#1#2{\partialdevdev{#1}{#2}}
 \global\long\def\partialdevdev#1#2{\frac{\partial^{2}#1}{\partial#2\partial#2^{\top}}}

\global\long\def\closure{\text{cl}}

\global\long\def\cpr#1#2{\Pr\left(#1\ |\ #2\right)}

\global\long\def\var{\text{Var}}

\global\long\def\Var#1{\text{Var}\left[#1\right]}

\global\long\def\cov{\text{Cov}}

\global\long\def\Cov#1{\cov\left[ #1 \right]}

\global\long\def\COV#1#2{\underset{#2}{\cov}\left[ #1 \right]}

\global\long\def\corr{\text{Corr}}

\global\long\def\sst{\text{T}}

\global\long\def\SST{\sst}

\global\long\def\ess{\mathbb{E}}

\global\long\def\Ess#1{\ess\left[#1\right]}

\newcommandx\ESS[2][usedefault, addprefix=\global, 1=]{\underset{#2}{\ess}\left[#1\right]}

\global\long\def\fisher{\mathcal{F}}

\global\long\def\bfield{\mathcal{B}}
 \global\long\def\borel{\mathcal{B}}

\global\long\def\bernpdf{\text{Bernoulli}}

\global\long\def\betapdf{\text{Beta}}

\global\long\def\dirpdf{\text{Dir}}

\global\long\def\gammapdf{\text{Gamma}}

\global\long\def\gaussden#1#2{\text{Normal}\left(#1, #2 \right) }

\global\long\def\gauss{\mathbf{N}}

\global\long\def\gausspdf#1#2#3{\text{Normal}\left( #1 \lcabra{#2, #3}\right) }

\global\long\def\multpdf{\text{Mult}}

\global\long\def\poiss{\text{Pois}}

\global\long\def\poissonpdf{\text{Poisson}}

\global\long\def\pgpdf{\text{PG}}

\global\long\def\wshpdf{\text{Wish}}

\global\long\def\iwshpdf{\text{InvWish}}

\global\long\def\nwpdf{\text{NW}}

\global\long\def\niwpdf{\text{NIW}}

\global\long\def\studentpdf{\text{Student}}

\global\long\def\unipdf{\text{Uni}}

\global\long\def\transp#1{\transpose{#1}}
 \global\long\def\transpose#1{#1^{\mathsf{T}}}

\global\long\def\mgt{\succ}

\global\long\def\mge{\succeq}

\global\long\def\idenmat{\mathbf{I}}

\global\long\def\trace{\mathrm{tr}}

\global\long\def\argmax#1{\underset{_{#1}}{\text{argmax}} }

\global\long\def\argmin#1{\underset{_{#1}}{\text{argmin}\ } }

\global\long\def\diag{\text{diag}}

\global\long\def\norm{}

\global\long\def\spn{\text{span}}

\global\long\def\vtspace{\mathcal{V}}

\global\long\def\field{\mathcal{F}}
 \global\long\def\ffield{\mathcal{F}}

\global\long\def\inner#1#2{\left\langle #1,#2\right\rangle }
 \global\long\def\iprod#1#2{\inner{#1}{#2}}

\global\long\def\dprod#1#2{#1 \cdot#2}

\global\long\def\norm#1{\left\Vert #1\right\Vert }

\global\long\def\entro{\mathbb{H}}

\global\long\def\entropy{\mathbb{H}}

\global\long\def\Entro#1{\entro\left[#1\right]}

\global\long\def\Entropy#1{\Entro{#1}}

\global\long\def\mutinfo{\mathbb{I}}

\global\long\def\relH{\mathit{D}}

\global\long\def\reldiv#1#2{\relH\left(#1||#2\right)}

\global\long\def\KL{KL}

\global\long\def\KLdiv#1#2{\KL\left(#1\parallel#2\right)}
 \global\long\def\KLdivergence#1#2{\KL\left(#1\ \parallel\ #2\right)}

\global\long\def\crossH{\mathcal{C}}
 \global\long\def\crossentropy{\mathcal{C}}

\global\long\def\crossHxy#1#2{\crossentropy\left(#1\parallel#2\right)}

\global\long\def\breg{\text{BD}}

\global\long\def\lcabra#1{\left|#1\right.}

\global\long\def\lbra#1{\lcabra{#1}}

\global\long\def\rcabra#1{\left.#1\right|}

\global\long\def\rbra#1{\rcabra{#1}}